\documentclass{article}
\usepackage{ijcai21}

\usepackage{graphics} % for pdf, bitmapped graphics files
\usepackage{epsfig} % for postscript graphics files
\usepackage{mathptmx} % assumes new font selection scheme installed
\usepackage{times} % assumes new font selection scheme installed
\usepackage{amsmath} % assumes amsmath package installed
\usepackage{amssymb}  % assumes amsmath package installed
\usepackage{cite}
\usepackage{amsmath,amssymb,amsfonts}
\usepackage{algorithmic}
\usepackage{graphicx}
\usepackage{caption}
\usepackage{subcaption}
\usepackage{textcomp}
\usepackage{xspace}
\usepackage{balance}
\usepackage{paralist}
\usepackage{url}
\usepackage[draft]{hyperref}
\usepackage[dvipsnames]{xcolor}
\usepackage{multirow}
\usepackage{subcaption}
\usepackage{tabularx}
\usepackage{multirow}
\usepackage{dsfont}
\usepackage{booktabs}
\usepackage{ textcomp }
\usepackage{blindtext}
\title{\LARGE \bf
Towards a Safety Case for Hardware Fault Tolerance in Convolutional Neural Networks Using Activation Range Supervision
}

\graphicspath{{./pics/}}

\author{
Florian Geissler$^1$\footnote{corresponding author, Email: florian.geissler@intel.com. Copyright © 2021 for this paper by its authors. Use permitted under Creative Commons License Attribution 4.0 International (CC BY 4.0).}\and
Syed Qutub$^1$\and
Sayanta Roychowdhury$^1$\and
Ali Asgari$^2$ \and
Yang Peng$^1$ \and
Akash Dhamasia$^1$ \and
Ralf Graefe$^1$ \and
Karthik Pattabiraman$^2$ \And
Michael Paulitsch$^1$
\affiliations
$^1$Intel, Germany\\
$^2$University of British Columbia, Canada
}

\newcommand{\oob}{\text{oob}}

\newcommand{\sdc}{\text{sdc}}
\newcommand{\due}{\text{due}}
\newcommand{\cl}{\text{cl}}
\newcommand{\ib}{\text{ib}}
\newcommand{\tp}{\text{Tp}}
\newcommand{\fp}{\text{Fp}}
\newcommand{\fn}{\text{Fn}}

\newcommand{\msb}{\text{MSB}}

\newcommand{\Tup}{T_{\text{up}}}
\newcommand{\Tlow}{T_{\text{low}}}

\newcommand{\find}{\text{ind}}
\newcommand{\favg}{f_{\text{avg}}}
\newcommand{\Cout}{C_{\text{out}}}

\newcommand{\treg}{${}^{\text{\textregistered}}$}
\newcommand{\ttm}{${}^{\text{\texttrademark}}$}

\begin{document}

\maketitle

%TODO: 
% - (TODO: accuracy is not safety)
% - (TODO: not every msb is mcl but almost every mcl is msb)
% - (TODO: discuss safety for CNN sota? )
% - (TODO: other methods dont show benefit yet since faults are mostly in later layers (have more parameters))
% - (TODO: overhead mention somewhere (conclusion)?)
% - shorten

%General:
%-: smaller weights mean high resilience?
%-: early layers more sensitive to weight faults, later layers more sensitive to neuron faults
%-: move to higher fault rates as well (20)?
% -: modify estimate 1/32, not every MSB flip leads to SDC but every SDC comes from MSB flips. Should be between 1/32 and 1/32*1/2
%Miovision data set, ResNet50 as a use case. Define safety-critical class confusions and uncritical ones.
%\begin{itemize}
%\item Safety-critical confusions wrt bound variation:
%Do the safety-critical confusions depend a lot on the extracted bounds? Vary by a few percent up, down 
%\item Safety-critical confusions wrt range restriction method:
%What ranger alternatives handle safety-critical confusions best? Can we define an optimum?
%\end{itemize}

%%%%%%%%%%%%%%%%%%%%%%%%%%%%%%%%%%%%%%%%%%%%%%%%%%%%%%%%%%%%%%%%%%%%%%%%%%%%%%%%
\begin{abstract}
Convolutional neural networks (CNNs) have become an established part of numerous safety-critical computer vision applications, including human robot interactions and automated driving.
Real-world implementations will need to guarantee their  robustness against hardware soft errors corrupting the underlying platform memory. Based on the previously observed efficacy of activation clipping techniques, we build a prototypical safety case for classifier CNNs by demonstrating that range supervision represents a highly reliable fault detector and mitigator with respect to relevant bit flips, adopting an eight-exponent floating point data representation. We further explore novel, non-uniform range restriction methods that effectively suppress the probability of silent data corruptions and uncorrectable errors. As a safety-relevant end-to-end use case, we showcase the benefit of our approach in a vehicle classification scenario, using ResNet-50 and the traffic camera data set MIOVision. The quantitative evidence provided in this work can be leveraged to inspire further and possibly more complex CNN safety arguments.
\end{abstract}

%%%%%%%%%%%%%%%%%%%%%%%%%%%%%%%%%%%%%%%%%%%%%%%%%%%%%%%%%%%%%%%%%%%%%%%%%%%%%%%%

\section{Motivation}
\label{sec:intro}

% Motive the problem of soft errors
With the widespread use of convolutional neural networks (CNN) across many safety-critical domains such as automated robots and cars, one of the most prevailing challenges is the establishment of a safety certification for such artificial intelligence (AI) components, e.g., with respect to the ISO 26262 \cite{Iso26262} or ISO/PAS 21448 (SOTIF) \cite{Sotif2019}. 
This certification requires not only a high fault-tolerance of the trained network against unknown or adversarial input, but also efficient protection against hardware faults of the underlying platform \cite{Athavale2020, Dixit2021}. Importantly this includes transient \textit{soft errors}, meaning disturbances originating from events such as cosmic neutron radiation, isotopes emitting alpha particles, or electromagnetic leakage on the computer circuitry itself. 

Soft errors typically manifest as single or multiple bit upsets in the platform's memory elements \cite{Neale2016}. As a consequence, network parameters (\textit{weight faults}) or local computational states (\textit{neuron faults}) can be altered during inference time, and invalidate the network prediction in a safety-critical way, for example, by misclassifying a person as a background image in an automated driving context \cite{Li2017, hoang2019ftclipact, chen2020ranger}. 
This has led to a search for strategies to verify CNN-based systems against hardware faults at the inference stage~\cite{Cluzeau2020}.
%Therefore, run-time monitoring methods have been identified as a strategy to verify learning systems against hardware faults at the inference stage \cite{Cluzeau2020}.
With chip technology nodes scaling to smaller sizes and larger memory density per area, future platforms are expected to be even more susceptible to soft errors \cite{Neale2016}.

\begin{figure}[t]
\centering
\includegraphics[width=.5\textwidth]{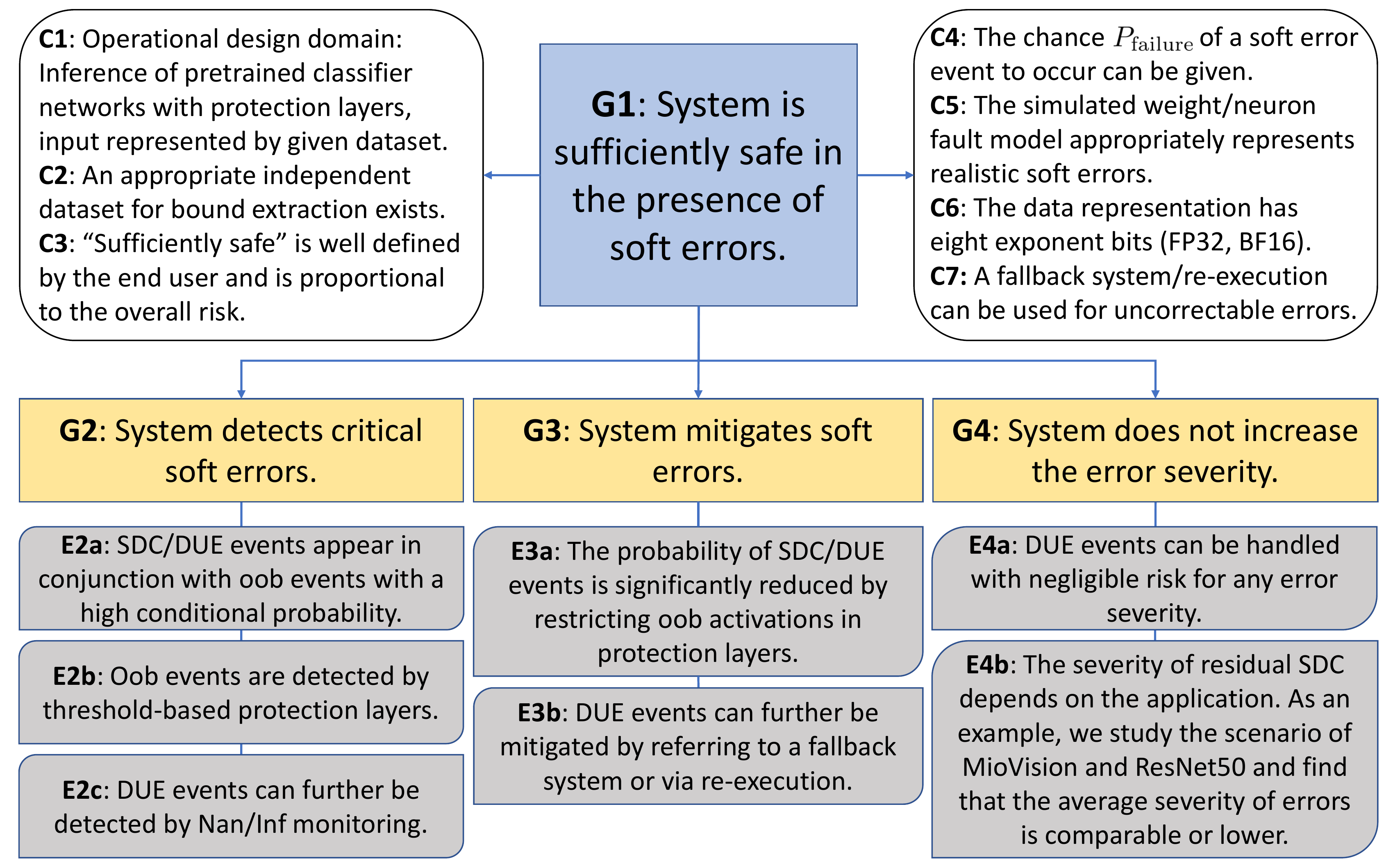}
\caption{Structured safety argument for the fault tolerance of a CNN in the presence of soft errors, using range restrictions. The notation follows \cite{Koopman2019} including goals (G), context (C), and evidence (E). "Oob" denotes "out-of-bounds".}
\label{fig:safety_case}
%\vspace{-0.5cm}
\end{figure}

\begin{figure}[t]
\centering
\includegraphics[width=.5\textwidth]{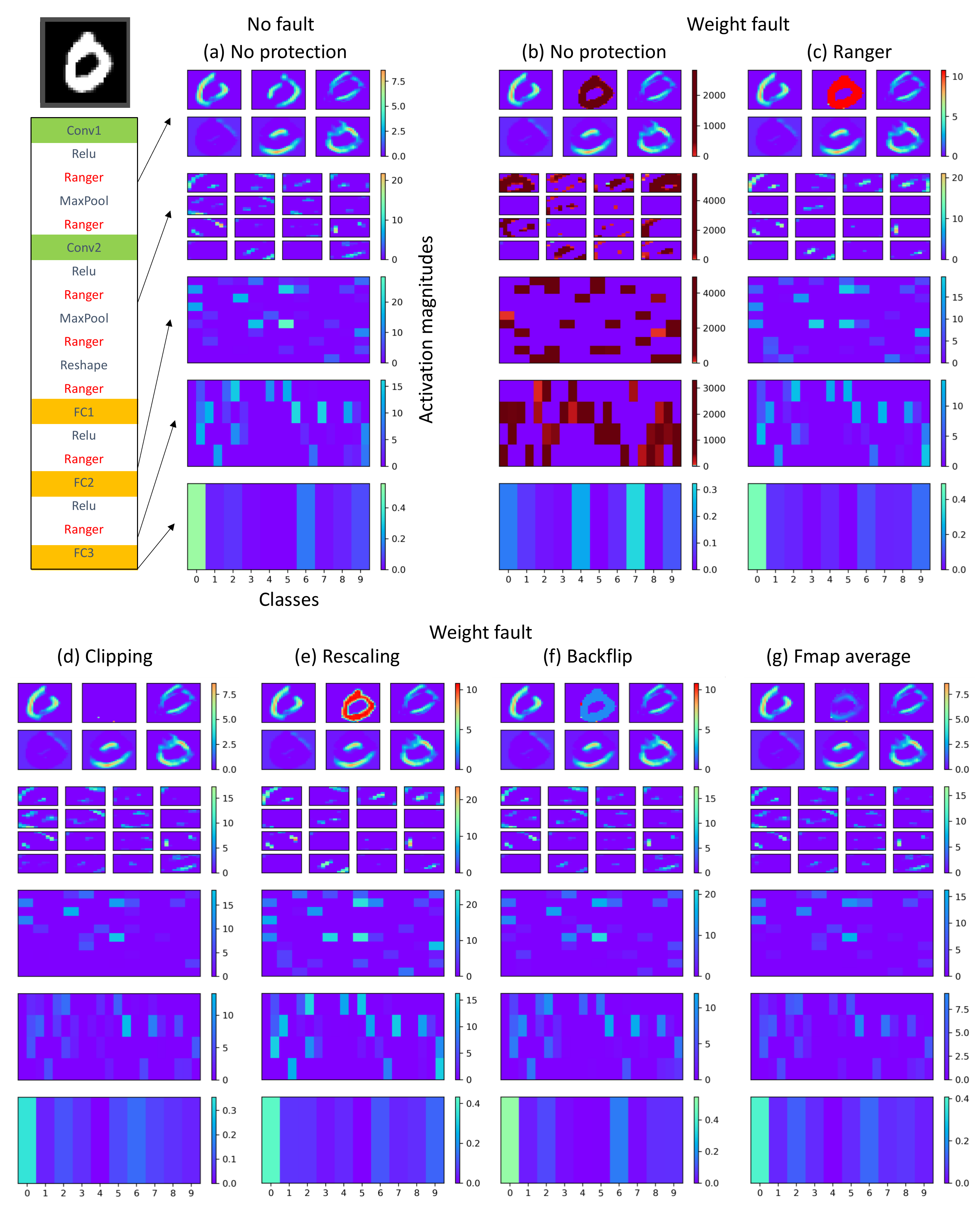}
\caption{Visualization example of the impact of a weight fault using LeNet-5 and the MNIST data set. Range restriction layers ("Ranger") are inserted following \cite{chen2020ranger} (top left). The rows represent the feature maps of the individual network layers after range restriction was applied, where linear layers (FC1-FC3) were reshaped to a 2D feature map as well for visualization purposes. In (b) - (g), a large weight fault value is injected in the second filter of the first convolutional layer. For the unprotected model (b), this leads to a SDC event ("0" gets changed to "7"). The columns (c) - (g) then illustrate the effect of the different investigated range restriction methods.}
\label{fig:lenet_visuals}
%\vspace{-0.5cm}
\end{figure}

In this paper, we evaluate range restriction techniques in CNNs exposed to platform soft errors with respect to the key elements of a prototypical safety case. 
This means that we formulate arguments (in the form of "goals") that constitute essential parts of a complete safety case, and provide quantitative evidence to support these goals in the studied context (see Fig.~\ref{fig:safety_case}). 
Individual safety arguments can be reused as building blocks of more complex safety cases.
The structure of our goals is based on the probabilistic, high-level safety objective of minimizing the overall risk \cite{Koopman2019}, expressed as: 
\begin{equation}
\label{eq:risk}
	\begin{aligned}
		P_{\text{loss}}(i) & = P_{\text{failure}}(i) \left[ (1-P_{\text{detection}}(i)) + (1-P_{\text{mitigation}}(i))\right], \\
		\text{Risk} & = \sum_i P_{\text{loss}}(i) \cdot \text{Severity}(i).
	\end{aligned}
\end{equation}
Explicitly, for a fault type $i$, this includes the sub-goals of efficient error \textit{detection} and \textit{mitigation}, as well as a consideration of the fault \textit{severity} in a given use case. 
%Our analysis provides quantitative evidence to support the above goals in the selected setups, see Fig.~\ref{fig:safety_case}. 
%Here, $i$ denotes a given fault type
%which most importantly includes efficient error \textit{detection} and \textit{mitigation}, as well as a consideration of the fault \textit{severity} for a given use case. 
%Explicitly, without human supervision, we have for a fault type $i$ \cite{Koopman2019},
On the other hand, the probability of occurrence of a soft error (i.e., $P_{\text{failure}}$ in Eq.~\ref{eq:risk}) is assumed to be a constant system property that cannot be controlled by run-time monitoring methods such as activation range supervision.
% --------------

In a nutshell, range restriction builds on the observation that silent data corruption (SDC) and detected uncorrectable errors (DUE), e.g., \textit{NaN} and \textit{Inf} occurrences), stem primarily from those bit flips that cause very large values, for example in high exponential bits \cite{Li2017}. Those events result in large activation peaks that typically grow even more during forward propagation due to the monotonicity of most neural network operations \cite{Chen2019a}. 
%We verify this effect by quantifying the correlation between out-of-bound and SDC/DUE events in Sec.~\ref{sec:error_detection}. 
To suppress the propagation of such corrupted values, additional range restriction layers are inserted in the network at strategic positions following the approach of Chen et. al.~\cite{chen2020ranger} (see Fig.~\ref{fig:lenet_visuals} for an example).
At inference time, the protection layers then compare the intermediate activations against previously extracted interval thresholds in order to detect and reset anomalously large values.
Derivative approaches have been shown to be efficient in recovering network performance \cite{Li2017, Hong2019, chen2020ranger, hoang2019ftclipact} and, advantageously, do not require the retraining of CNN parameters nor computationally expensive functional duplications.

The focus of this paper is to examine alternative restriction schemes for optimized soft error mitigation.
In a CNN, the output of every kernel is represented as a two-dimensional (2D) \textit{feature map}, where the activation magnitudes encode specific features, on which the network bases its prediction. 
Soft errors will manifest as distortions of feature maps in all subsequent layers that make use of the corrupted value, as shown in Fig.~\ref{fig:lenet_visuals}(a)-(b). The problem of mitigating soft errors in a CNN can therefore be rephrased as restoring the fault-free topology of feature maps.
 %and in particular to suppress unnaturally large activation peaks.
%A focus of this paper is to examine restriction schemes for optimal error mitigation.

Previous analyses have adopted uniform range restriction schemes that truncate out-of-bound values to a finite threshold \cite{chen2020ranger, hoang2019ftclipact}, e.g., Fig.~\ref{fig:lenet_visuals}(c)-(d). 
We instead follow the intuition that optimized, non-uniform range restriction methods that attempt to  reconstruct feature maps (see Fig.~\ref{fig:lenet_visuals}(e)-(g), and details in Sec.~\ref{sec:ranger_alternatives}) can not only reduce SDC to a comparable or even lower level, but may also lead to less critical misclassifications in the case of an SDC. This is because classes with more similar attributes will display more similar high-level features (e.g., pedestrian and biker will both exhibit upright silhouette, in contrast to car and truck classes).

Finally, a safety analysis has to consider that not all SDC events pose an equal risk to the user. We study a safety-critical use case evaluating cluster-wise class confusions in a vehicle classification scenario (Sec.~\ref{sec:safety}). The example shows that range supervision reduces the severe confusions proportionally with the overall number of confusions, meaning that the total risk is indeed mitigated.  

In summary, this paper make the following contributions:
\begin{itemize}
\item \textbf{Fault detection:} We quantify the correlation between SDC events and the occurrence of out-of-bound activations to demonstrate the high efficiency of fault detection by monitoring intermediate activations,
\item \textbf{Fault mitigation:} We explore three novel range restriction methods that build on the preservation of the feature map topologies instead of mere value truncation,
\item \textbf{Fault severity:} We demonstrate the benefit of range supervision in an end-to-end use case of vehicle classification where high and low severities are estimated by the generic safety-criticality of class confusions.
\end{itemize}
The article is structured as follows: Section~\ref{sec:sota} reviews relevant previous work while section~\ref{sec:setup} describes the setup used in this paper.
Subsequently, the sections~\ref{sec:error_detection}, \ref{sec:ranger_alternatives}, and \ref{sec:safety} discuss error detection, mitigation, and an exemplary risk analysis, respectively, before section~\ref{sec:conclusion} concludes the paper.

%%%%%%%%%%%%%%%%%%%%%%%%%%%%%%%%%%%%%%%%%%%%%%%%%%%%%%%%%%%%%%%%%%%%%%%%%%%%%%%%

\section{Related work}
\label{sec:sota}
Parity or error-correcting code (ECC) can protect memory elements against single soft errors \cite{Neale2016, Lotfi2019}. However, due to the high compute and area overhead, this is typically done only for selected critical memory blocks.
Component replication techniques such as triple modular redundancy can be used for the full CNN execution at the cost of a large overhead.
Selective hardening of hardware elements with the most salient parameters can improve the robustness of program execution in the presence of underlying faults \cite{Li2017, Hanif2020}.
On a software level, the estimation of the CNN's vulnerable feature maps (fmaps) and the selective protection by duplicated computations \cite{mahmoud2020hardnn}, or the assertive re-execution with stored, healthy reference values \cite{Ponader2020} has been investigated. Approaches using algorithm-based fault tolerance (ABFT) \cite{Zhao2021} seek to protect networks against soft errors by checking invariants that are characteristic for a specific operation (e.g., matrix multiplication). Symptom-based error detection may for example include the interpretation of feature map traces by a secondary companion network \cite{Schorn2018}. The restriction of intermediate ranges was explored \cite{Li2017, Hong2019} in the form of modified (layer-insensitive) activation functions such as $tanh$ or $ReLU6$. This concept was extended to find specific uniform protection thresholds for neuron faults \cite{chen2020ranger} or clipping bounds for weight faults \cite{hoang2019ftclipact}.  An alternative line of research is centered around fault-aware retraining~\cite{Yang2017}.

%%%%%%%%%%%%%%%%%%%%%%%%%%%%%%%%%%%%%%%%%%%%%%%%%%%%%%%%%%%%%%%%%%%%%%%%%%%%%%%%

\section{Experimental setup}
\label{sec:setup}
\subsection{Models, data sets, and system}
CNNs are the most commonly used network variant for computer vision tasks such as object classification and detection.
We compare the three standard classifier CNNs ResNet-50 \cite{He2016}, VGG-16 \cite{Simonyan2015}, and AlexNet \cite{Krizhevsky2012} together with the test data set ImageNet \cite{JiaDeng2009} and MIOVision \cite{Luo2018} for the investigation of a safety-critical example use case.
Since fault injection is compute-intensive, we rescale our test data set for ImageNet to a subset of $1000$ images representing $20$ randomly selected classes. For MIOVision, a subset of $1100$ images ($100$ per class) that were correctly classified in the absence of faults was chosen.
All experiments adopt a single-precision floating point format (FP32) according to the IEEE754 standard \cite{IEEE2019}. Our conclusions apply as well to other floating point formats with the same number of exponent bits, such as  BF16 \cite{Intel2018}, since no relevant effect was observed from fault injections in mantissa bits (Sec.~\ref{sec:error_detection}).

Experiments were performed in PyTorch (version 1.8.0) deploying torchvision models (version 0.9.0). For MIOVision, the ResNet-50 model was retrained \cite{Theagarajan2017}.
We used Intel\treg\ Core\ttm\ i9 CPUs, with inferences running on GeForce RTX $2080$, Titan RTX, and RTX 3090 GPUs.

\subsection{Protection layers and bound extraction}
We insert protection layers at strategic positions in the network such as after activation, pooling, reshape or concatenate layers, according to the model of Chen et al. \cite{chen2020ranger}.
Each protection layer requires specific bound values for the expected activation ranges as a parameter. We extract those by monitoring the minimal and maximal activations from a separate test input, which is taken from the training data sets of ImageNet ($143K$ images used) and MIOVision ($83K$ images used), respectively. This step has to be performed only once. Bound extraction depends on the data set and will in general impact the safety argument (see Fig.~\ref{fig:safety_case}).
To check the suitability of the  bounds, we verify that no out-of-bound events were detected during the test phase in the absence of faults, so the baseline accuracy is the same with and without protection. 
%To give an overview, 
While all minimum bounds are zero in the studied setup, the maximum activation values for ImageNet vary by layer in a range of (see also Sec.~\ref{sec:ranger_alternatives}) $1 < \Tup < 45$ for ResNet-50, $20 < \Tup < 360$ for VGG-16, and $65 < \Tup < 170$ for AlexNet. For MIOVision and ResNet-50, we find maximum bounds between $1 < \Tup < 19$.
%
%Note that bound extraction represents a constraint of the framework, as a separate dataset is required for this purpose.
% 550*1300 Images in traning data set randomly selected, used 20%*550*1300 = 143K images.
% MIOvision: 20%  * 415330  = 83K randomly selected

\begin{figure}[t]
\centering
\includegraphics[width=.48\textwidth]{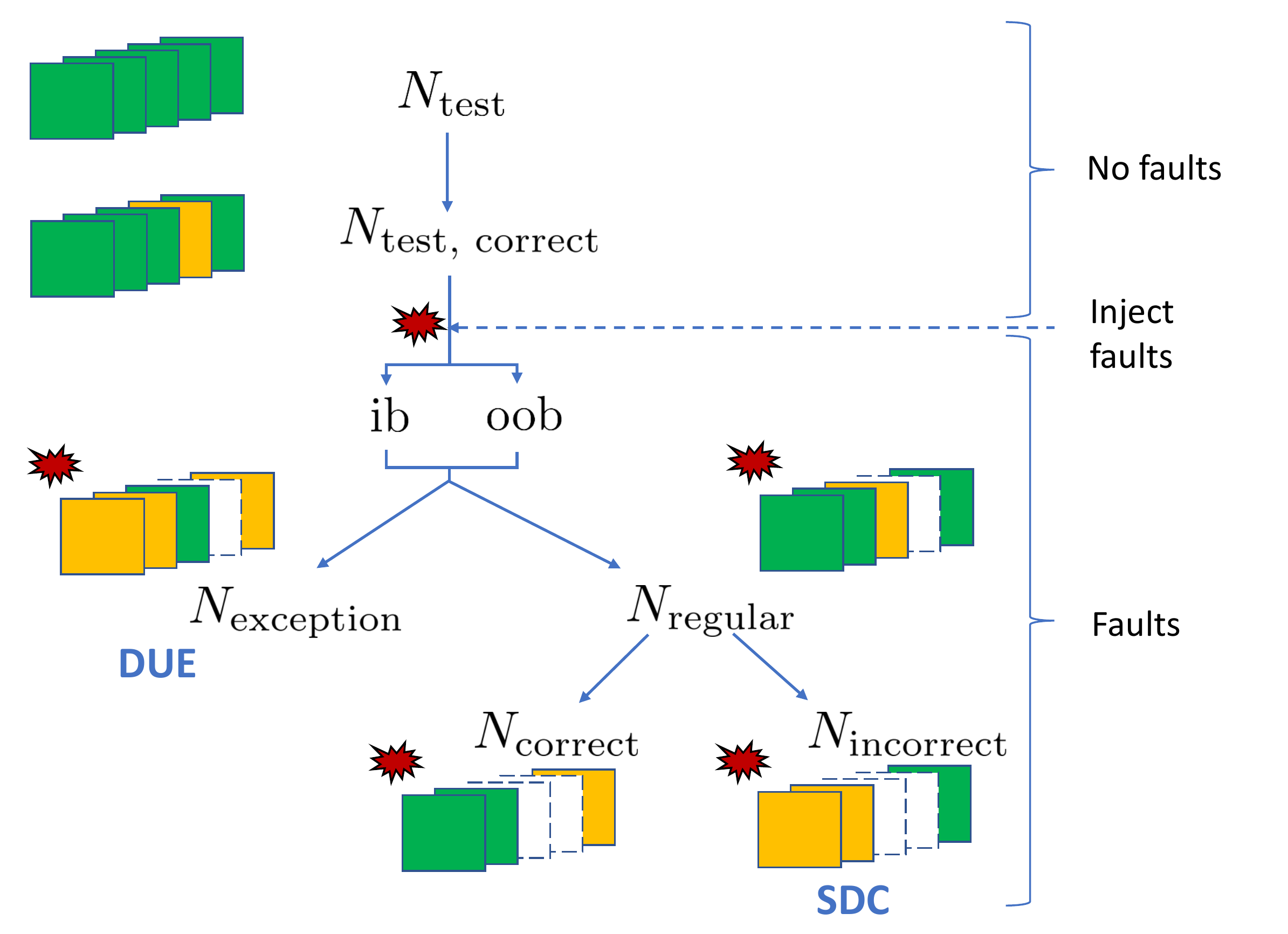}
\caption{Illustration of SDC and DUE events. Errors are detected or missed in the case of out-of-bound (oob) or in-bound (ib) events, respectively. (Green) Samples of the data set that form the subset of a given filtering stage, (Yellow) samples of the data set that are discarded at the given stage, (White) samples that were filtered out at a previous stage.
%$N_{\text{test}}$ is the number of images in the initial test data set which gets filtered ($N_{\text{test, correct}}$) to exclude input that is misclassified already in the absence of faults. After fault injection, range supervision can detect (out-of-bound event) the error or not (in-bound event). The number of images with and without exceptions is given by $N_{\text{exception}}$ and $N_{\text{regular}}$, respectively, where the former represent DUE events. Regular input can be classified correctly ($N_{\text{correct}}$) or incorrectly ($N_{\text{incorrect}}$, leading to a SDC event).
}
\label{fig:sdc_due}
%\vspace{-0.5cm}
\end{figure}

\subsection{Fault model and injection}

% Discuss types of soft errors (weights, neurons)
In line with previous investigations, we distinguish two different manifestations of memory bit flips referred to here as \textit{weight faults} and \textit{neuron faults}. 
The former represent soft errors affecting memory elements that store the learned network parameters, while the latter refer to errors in memory that holds temporary states such as intermediate network layer outputs.
While neuron faults may also impact states used for logical instructions, it was demonstrated that bit flip injections in the output of the affected layer are generally a good model approximation \cite{Chang2019}.
Memory elements can be protected against single bit flips by mechanisms such as parity and ECC \cite{Neale2016, Lotfi2019}. However, this kind of protection is not always available due to the associated compute and area overhead. Further, ECC typically cannot correct multi-bit flips. 
%Also, even occasional weight faults bear a high risk due to the frequent reuse of the parameter during a single forward pass. On the other hand, neuron faults represent faults that occur in the processor's data path such as arithmetic logic units and pipeline registers \cite{Li2018a, chen2020ranger}. Even though those type of soft errors may take various forms at an instruction level, it was demonstrated that bit flip injections in the output of the affected layer are a good model approximation \cite{Chang2019}.

We inject faults either directly in the weights of CNN layers (\textit{weight faults}) or in the output of the latter (\textit{neuron faults}), using a customized fault injection framework based on PytorchFI \cite{Mahmoud2020a}. To speed up the experiments we focus on bit flips in the most relevant bit positions $0-8$ (sign bit and exponential bits, neglecting mantissa) unless stated otherwise. Fault locations (i.e., layer index, kernel index, channel etc.) in the network are randomly chosen with an equal weight, so without further constraints on the selection process to reflect the arbitrary occurrence of soft errors.
%
%A slightly different injection logic is adopted for weight and neuron faults: 
As weights are typically stored in the main memory and loaded only once for a given application, we keep the same weight faults for one entire epoch, running all tested input images. In total, we run $500$ epochs, i.e., fault configurations, each one applied to $1K$ images.
%This means the number of different faults will be equivalent to the number of epochs times the number of faults per image. 
Neuron faults, on the other hand, apply to memory representing temporary states that are overwritten for each new input. Therefore, we inject new neuron faults for each new input and run $100$ epochs resulting in $100K$ fault configurations, each one applied to a single image.
 %In all experiments we run $500$ epochs for weight, and $100$ epochs for neuron faults.
%, such that the number of injected faults equals the number of input images times the epochs times the number of faults per image.
%NOTE: all errors denote the 95% confidence interval

\subsection{Evaluation}
To quantify the impact of faults on the system safety, we measure the rate of SDC events. Throughout, we consider the Top-1 prediction to determine SDC. In line with previous work \cite{chen2020ranger, Li2017}, SDC is defined as the ratio of images that are misclassified in the presence of faults (without exceptions) but correctly classified in the absence of faults and the overall number of images, $p(\sdc)=N_{\text{incorrect}}/N_{\text{test , correct}}$ (Fig.~\ref{fig:sdc_due}). 
%In contrast to the accuracy metric, the SDC thus rules out the imperfect classification capacity of the network without faults. 

During the forward pass, non-numerical exceptions in the form of \textit{Inf} and \textit{NaN} values can be encountered, due to the following reasons: \textbf{i)} \textit{Inf} values occur if large activation values accumulate (for example during conv2d, linear, avgpool2d operations) until they exceed the maximum of the data representation. This effect becomes particularly apparent when flips of the most significant bit (MSB, position index $1$) are injected. \textbf{ii)} \textit{NaN} values are found when denominators are undetermined or multiple \textit{Inf} values get added, e.g., in BatchNorm2d layers, \textbf{iii)} \textit{NaN} values can be generated directly via bit flips in conv2d layers due to the fact that FP32 encodes \textit{NaN} as all eight exponent bits being in state $"1"$. In the studied classifier networks, the latter effect is very rare for single bit flips in weights (see Sec.~\ref{sec:error_detection}) but not necessarily for single neuron bit flips or multiple flips of either type.

The creation of the above exceptions is found to differ slightly between CPU and GPU executions, as well as between experiments with different batch sizes on the accelerator. We attribute this observation to algorithmic optimizations on the GPU that are not necessarily IEEE754-compliant and thus affect the floating point precision \cite{nvidia2021}.
To mitigate the effect of exception handling we monitor the occurrences of \textit{Inf} and \textit{NaN} in the output of any network layer. All forward passes with an exception are separated and define the detected uncorrectable error (DUE) rate, $p(\due)=N_{\text{exceptions}}/N_{\text{test, correct}}$, see Fig.~\ref{fig:sdc_due}.

In a real system, DUE events can be readily monitored and the execution is typically halted on detection. However, due to the non-numerical nature of these errors we cannot apply the same mitigation strategy that is adopted for SDC events. We therefore make the assumption that either a fallback system (e.g., alternative classifier, emergency stop of vehicle, etc.) can be leveraged or a timely re-execution is possible to recover from transient DUE events. This in turn assumes that DUEs do not impact the system safety but may compromise the system availability when occurring frequently.

%%%%%%%%%%%%%%%%%%%%%%%%%%%%%%%%%%%%%%%%%%%%%%%%%%%%%%%%%%%%%%%%%%%%%%%%%%%%%%%%

\section{Error detection coverage}
\label{sec:error_detection}

\begin{figure}[t]
\centering
\includegraphics[width=.5\textwidth]{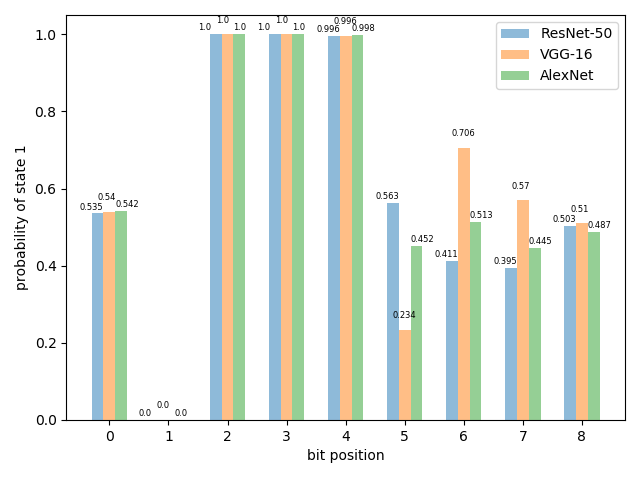} %[width=0.5\textwidth]
\caption{Bit-distribution across all weight parameters in conv2d layers. Values are represented in FP32, where only the sign bit ($0$) and the exponent bits ($1-8$) are shown.}
\label{fig:weight_dist}
%\vspace{-0.5cm}
\end{figure}

To effectively protect the network against faults, we first verify the error detection coverage for silent errors.
Those errors are detected by a given protection layer if the activation values exceed (fall short of) the upper (lower) bound.
If at least one protection layer is triggered per inference run, we register an out-of-bound ($\oob$) event. Otherwise, we have an in-bound ($\ib$) event.
In addition, we quantify the probabilities of SDC and regular correct classification $(\cl)$ events, as well as the respective conditional probabilities that correct and incorrect classifications occur \textit{given that $\oob$ or $\ib$ events were detected}. This allows us to define true positive ($\tp$), false positive ($\fp$), and false negative ($\fn$) SDC detection rates as
\begin{equation}
\label{eq:tpfpfn}
	\begin{aligned}
	\tp & = p(\sdc | \oob)\cdot p(\oob), \\
	\fp & = p(\cl | \oob)\cdot p(\oob), \\
	\fn & = p(\sdc | \ib)\cdot p(\ib).
	\end{aligned}
\end{equation}
The fault detector then is characterized by precision, $P = \tp/(\tp + \fp)$, and recall, $R = \tp/(\tp + \fn)$.

The Tab.~\ref{tab:prec_rec} displays the chances of $\oob$ and $\sdc$ events resulting from a single fault per image in the absence of range protection.
For weight faults, we find that all three CNNs showcase a high correlation between $\oob$ situations and either SDC or DUE events ($p(\sdc|\oob) + p(\due|\oob)>0.99$), which can be associated with the chance of a successful error detection, $P_{\text{detection}}$ (see Eq.~\ref{eq:risk}). The chance of finding SDC after $\ib$ events is very small ($\ll 1e^{-3}$), leading to a very high precision and recall performance ($>0.99$).
%For weight faults, there is a probability of $>0.99$ that an $\oob$ situation also leads to SDC, while for $\ib$ events the equivalent chance is $<1e^{-5}$. 
%This leads to a very high precision and recall performance ($>0.99$) for weight fault detection. 
For neuron faults, while the recall remains very high, the precision is reduced (in particular VGG-16 and AlexNet) due to additional \fp\ events where non-MSB $\oob$ events still get classified correctly.

We further verify that SDC events from single weight faults are attributed almost exclusively to flips of the MSB. This can be explained with the distribution of parameters in the studied networks (Fig.~\ref{fig:weight_dist}). The weight values are closely centered around zero, and thus exhibit characteristic properties when represented in an eight-exponent data format. In the fault-free case, the MSB always has state ``0", while the exponent bits $2$ to $4$ are almost always in state ``1". This means that among the relevant exponential bits all single bit flips of the MSB will produce large values, while those of the other exponential bits will either be from $``1"\to ``0"$ or will be too small to have a significant effect. 
%From that observation we can reassure that SDC occurs statistically in $p(\mcl)\gtrsim 1/2\cdot 1/32 = 0.0156$ of all weight fault injections. 

For neuron faults, on the other hand, the distribution of fault-free values is input-dependent and broader, leading in general to a smaller quota of MSB flips to SDC, in favor of flips of other exponential bits and the sign bit. No SDC due to mantissa bit flips were observed in either  weight or neuron faults.
%
%and a slightly lower correlation of $\oob$ and $\mcl$ events.
%However, we still find high precisions ($\gtrsim 0.90$) and recalls ($\gtrsim 0.94$) in all three setups. 
%Here, flips of lower exponential bits and the sign bit contribute to a similar extent to the non-MSB SDC events. No SDC due to mantissa bit flips were observed in either of weight or neuron faults.
DUE events are unlikely ($<0.01$) for a single bit flip as there are not multiple large values to add up. Further, network weights are usually $<1$, meaning that at least two exponent bits are in state $"0"$, and hence at least two bit flips are needed to directly generate a \textit{NaN} value. 

\begin{table}[h!]
\centering
\begin{tabularx}{0.47\textwidth}{X X X} %{|p{1cm}|p{4cm}|p{5cm}|}
			\toprule
			 & Weight faults & Neuron faults \\ \midrule
			%1 & & & \\
			\textbf{ResNet-50}: & & \\
			$p(\sdc)$ & $0.018 \pm 0.001$ & $0.013 \pm 6e^{-4}$ \\ %\hline
			$p(\oob)$ & $0.019 \pm 0.001$ & $0.013 \pm 6e^{-4}$ \\ %\hline
			$p(\sdc | \oob)$ & $0.981 \pm 0.008$ & $0.974 \pm 0.008$\\ %\hline
			$p(\sdc | \ib)$ & $5e^{-5} \pm 4e^{-5}$ & $0.0 \pm 0.0$\\ %\hline
			$p(\msb|\sdc)$ & $0.998 \pm 0.002$ & $0.961 \pm 0.012$ \\
			P & $\mathbf{0.997 \pm 0.002}$ & $\mathbf{0.980 \pm 0.006}$ \\ %\hline
			R & $\mathbf{0.997 \pm 0.002}$ & $\mathbf{1.0 \pm 0.0}$\\ 
			$p(\due)$ & $3e^{-4} \pm 1e^{-4}$ & $5e^{-4} \pm 1e^{-4}$ \\ %\hline
			$p(\due | \oob)$ & $0.016 \pm 0.008$ & $0.006 \pm 0.005$\\ %\hline
			$p(\msb|\due)$ & $1.0 \pm 0.0$ & $1.0 \pm 0.0$ \\ \hline
			\textbf{VGG-16}: & & \\
			$p(\sdc)$ & $0.024 \pm 0.001$ & $0.016 \pm 9e^{-4}$ \\ %\hline
			$p(\oob)$ & $0.027 \pm 0.001$ & $0.020 \pm 0.001$ \\ %\hline
			$p(\sdc | \oob)$ & $0.893 \pm 0.010$ & $0.778 \pm 0.016$\\ %\hline
			$p(\sdc | \ib)$ & $7e^{-5} \pm 7e^{-5}$ & $0.0 \pm 0.0$\\ %\hline
			$p(\msb|\sdc)$ & $0.997 \pm 0.003$ & $0.397 \pm 0.017$ \\
			P & $\mathbf{0.999 \pm 0.001}$ & $\mathbf{0.820 \pm 0.014}$ \\ %\hline
			R & $\mathbf{0.997 \pm 0.003}$ & $\mathbf{1.0 \pm 0.0}$\\
			$p(\due)$ & $0.003 \pm 4e^{-4}$ & $0.006 \pm 4e^{-4}$ \\ %\hline
			$p(\due | \oob)$ & $0.106 \pm 0.011$ & $0.051 \pm 0.012$\\ %\hline
			$p(\msb|\due)$ & $1.0 \pm 0.0$ & $1.0 \pm 0.0$ \\ \hline
			\textbf{AlexNet}: & & \\
			$p(\sdc)$ & $0.022 \pm 0.001$ & $0.013 \pm 0.001$ \\ %\hline
			$p(\oob)$ & $0.024 \pm 0.001$ & $0.015 \pm 0.001$ \\ %\hline
			$p(\sdc | \oob)$ & $0.907 \pm 0.012$ & $0.877 \pm 0.023$\\ %\hline
			$p(\sdc | \ib)$ & $2e^{-4} \pm 1e^{-4}$ & $9e^{-5} \pm 5e^{-5}$\\ %\hline
			$p(\msb|\sdc)$ & $0.995 \pm 0.003$ & $0.245 \pm 0.031$ \\
			P & $\mathbf{1.0 \pm 0.0}$ & $\mathbf{0.913 \pm 0.025}$ \\ %\hline
			R & $\mathbf{0.989 \pm 0.005}$ & $\mathbf{0.994 \pm 0.004}$\\ %\hline
			$p(\due)$ & $0.003 \pm 3e^{-4}$ & $0.005 \pm 3e^{-4}$ \\ %\hline
			$p(\due | \oob)$ & $0.093 \pm 0.012$ & $0.040 \pm 0.011$\\ %\hline
			$p(\msb|\due)$ & $1.0 \pm 0.0$ & $1.0 \pm 0.0$ \\ \hline
			\bottomrule
\end{tabularx}
%\vspace{0.3cm}
\caption{Statistical absolute and conditional probabilities of SDC, DUE events and the related precision and recall of the fault detector. Experiments of $10K$ fault injections were repeated $10$ times, where a single fault per image was injected in any of the $32$ bits for each image (from ImageNet, using a batch size of one). We further list what proportion of SDC or DUE events were caused by MSB flips.}
\label{tab:prec_rec}
%\vspace{-0.3cm}
\end{table}

%%%%%%%%%%%%%%%%%%%%%%%%%%%%%%%%%%%%%%%%%%%%%%%%%%%%%%%%%%%%%%%%%%%%%%%%%%%%%%%%

\section{Range restriction methods for error mitigation}
\label{sec:ranger_alternatives}

\subsection{Model}
%A 2D convolutional layer maps an input tensor of dimensions $(N, \Cin, \Hin, \Win)$ to an output tensor of dimensions $(N, \Cout, \Hout, \Wout)$.
%Here, $N$ denotes the batch size, $\Cin$ ($\Cout$) the number of the input (output) channels, as well as $\Hin$ ($\Hout$) and $\Win$ ($\Wout$) the input (output) widths and heights, respectively.
%The number of output channels $\Cout$ equals the number of different filters that constitute the conv2d layer, and thus represents the features this layer seeks to analyze.
We refer to a subset of the tensor given by a specific index in the batch and channel dimensions as a 2D feature map, denoted by $f$. 
%
%A soft error in a conv2d corrupts one or multiple activations in the output fmaps. The task of mitigating this error can be considered as restoring the original (i.e., fault-free) topology of the individual feature maps. In the following, we start from two established restriction methods that seek to achieve this, and present additional novel extensions.
Let $x$ be an activation value from a given feature map tensor $f \in \{f_1, f_2, \ldots,  f_{\Cout}\}$. Further, $\Tup$ and $\Tlow$ denote the upper and lower activation bounds assigned to the protection layer, respectively.

% Ranger
\textbf{Ranger:} For a given set of $(f, \Tup, \Tlow)$, Ranger \cite{chen2020ranger} maps out-of-bound values to the expected interval (see Fig.~\ref{fig:lenet_visuals}c), 
\begin{equation}
r_{\text{ranger}}(x) =
\begin{cases}
\Tup & \text{if $x > \Tup$}, \\
\Tlow & \text{if $x < \Tlow$}, \\
x & \text{otherwise}.
\end{cases}
\label{eq:rranger}
\end{equation}

% Clipping
\textbf{Clipper:} In a similar way, clipping truncates activations that are out of bound to zero \cite{hoang2019ftclipact},
\begin{equation}
r_{\text{clipping}}(x) =
\begin{cases}
0 & \text{if $x > \Tup$ or $x < \Tlow$}, \\
x & \text{otherwise}.
\end{cases}
\label{eq:rclipping}
\end{equation}
The intuition is that it can be favorable to eliminate corrupted elements rather than to re-establish finite activations. 
%\ref{fig:lenet_visuals}

% Rescaling
\textbf{FmapRescale:} While uniform restriction methods help in eliminating large out-of-bound values, the information encoded in relative differences of activation magnitudes is lost when all out-of-bound values are flattened to the same value.
%Hence, we argue that non-uniform range restriction can lead to a more accurate recovery of the fmap topology, and thus to restored network performance.
The idea of rescaling is to linearly map all large out-of-bound values back onto the interval $[\Tlow, \Tup]$, implying that smaller out-of-bound values are reduced more. This follows the intuition that the out-of-bound values can originate from the entire spectrum of in-bound values. 
\begin{equation}
r_{\text{rescale}}(x) =
\begin{cases}
\frac{\left(x - \min(f)\right)\left(\Tup - \Tlow \right)}{\max(f) - \min(f)} + \Tlow & \text{if $x > \Tup$}, \\
\Tlow & \text{if $x < \Tlow$}, \\
x & \text{otherwise}.
\end{cases}
\label{eq:rrescale}
\end{equation}

% Backflip
\textbf{Backflip:} We analyze the underlying bit flips that may have caused out-of-bound values. This reasoning holds for neuronal faults, where we may assume that a specific activation value is bit-flipped directly. For weight faults, on the other hand, the observed out-of-bound output activation is the result of a multiply-and-accumulate operation of an input tensor with a bit-flipped weight value. However, we argue that the presented back-flip operation will recover a representative product, given that the input component is of the order of magnitude of one. To restore a flipped value, we distinguish the following cases:
\begin{equation}
r_{\text{backflip}}(x) =
\begin{cases}
0 & \text{if $x > \Tup\cdot 2^{64}$}, \\
2 & \text{if $\Tup\cdot 2^{64}> x > \Tup\cdot 2$}, \\
\Tup & \text{if $\Tup\cdot 2 > x > \Tup$}, \\
\Tlow & \text{if $x < \Tlow$}, \\
x & \text{otherwise}.
\end{cases}
\label{eq:rbackflip}
\end{equation}
The above thresholds are motivated by the following logic: Given appropriate bounds, an activation is $<\Tup$ before a bit flip. Any flip of an exponential bit $i \in \{1 \ldots 8 \}$ effectively multiplies a factor of $\text{pow}(2, 2^{8-i})$. Hence, any value beyond $\Tup\cdot 2^{64}$ must have originated from a flip $"0"\to "1"$ of the MSB, meaning that the original value was between $0$ and $2$. We then set back all out-of-bound values in this regime to zero, assuming that lower reset values represent a more conservative choice in eliminating faults.
Next, flipped values that are between $\Tup\cdot 2^{64}> x > \Tup\cdot 2$ can possibly originate from a flip of any exponential bit.
Given that $\Tup$ is typically $>1$, a bit flip has to produce a corrupted absolute value $>2$ in this regime. This is possible only if either the MSB is flipped from $"0"\to"1"$, or the MSB is already at $"1"$ and another exponential bit is flipped $"0"\to"1"$. In all variants of the latter case, the original value had to be already $>2$ itself, and hence we conservatively reset out-of-bound values to $2$.
Finally, corrupted values of $\Tup\cdot 2 > x > \Tup$ may originate from any non-sign bit flip. Lower exponential or even fraction bit flips result from already large values close to $\Tup$ in this case, which is why we set back those values to the upper bound. As in Ranger, values that are too small are reset to $\Tlow$.

\begin{figure}
     \centering
     \begin{subfigure}[htpb]{0.48\textwidth}
         \centering
         \includegraphics[width=\textwidth]{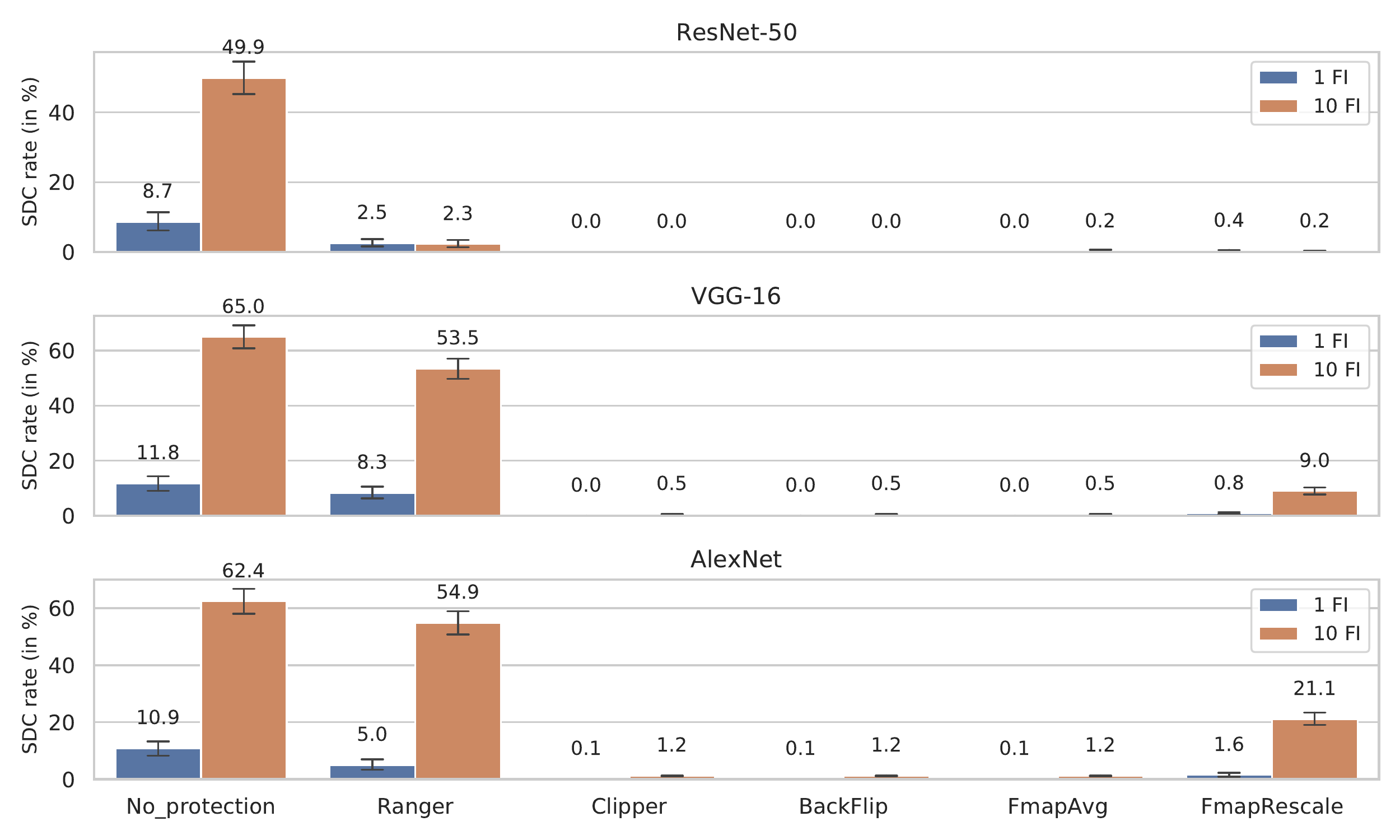}
         \caption{Weight faults}
         \label{fig:sdc_all_a}
     \end{subfigure} \\ %\hfill %\\
     \begin{subfigure}[t]{0.48\textwidth}
         \centering
         \includegraphics[width=\textwidth]{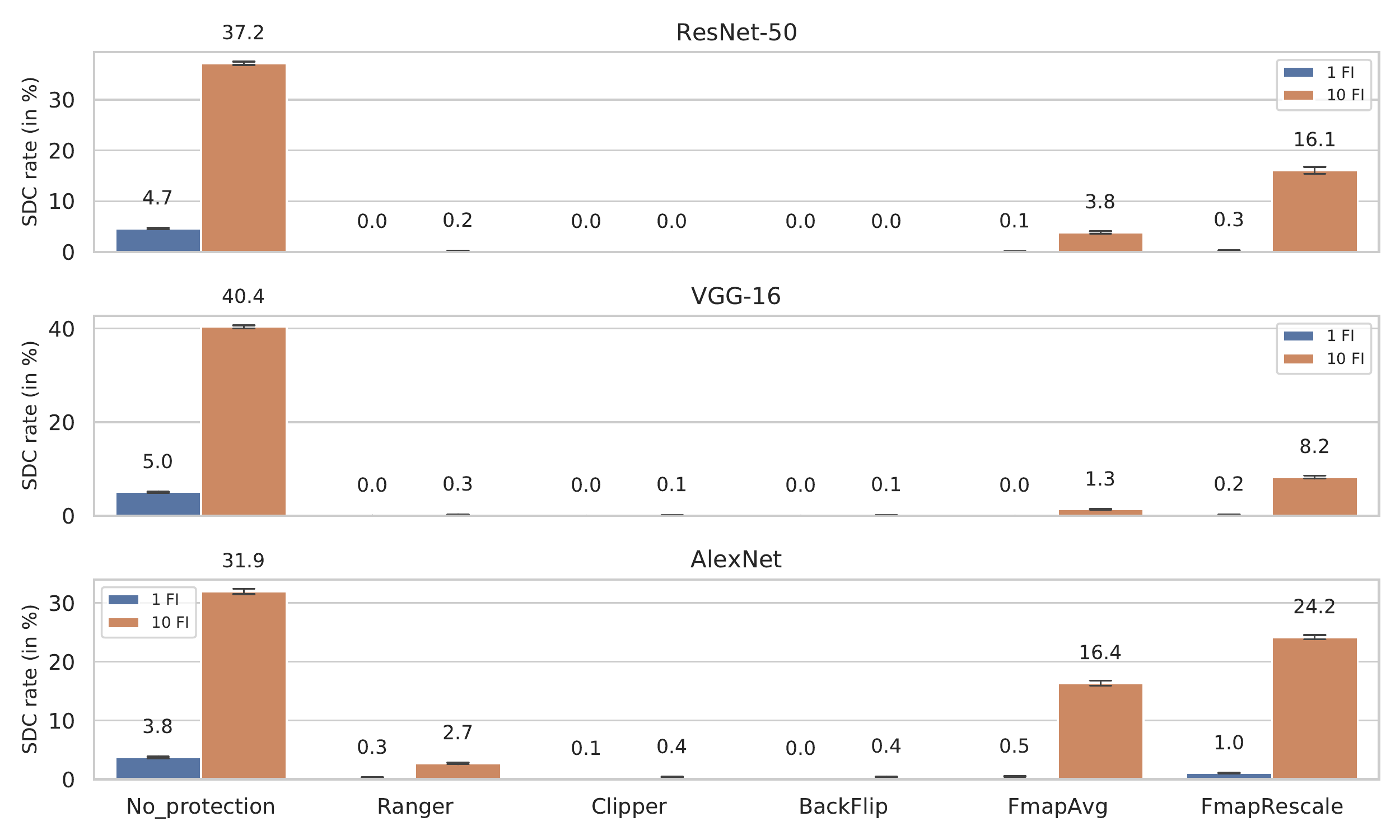}
         \caption{Neuron faults}
         \label{fig:sdc_all_b}
     \end{subfigure}
		\caption{SDC rates for weight (a) and neuron (b) faults using different range supervision techniques. Note that compared to Tab.~\ref{tab:prec_rec} rates are around $4\times$ higher since we inject only in the bits $0-8$ here.}
\label{fig:sdc_all}
\end{figure}

%Fmap avg
\textbf{FmapAvg:}
The last proposed range restriction technique uses the remaining, healthy fmaps of a convolutional layer to reconstruct a corrupted fmap. The intuition behind this approach is as follows: 
Every filter in a given conv2d layer tries to establish characteristic features of the input image. Typically, there is a certain redundancy in the topology of fmaps,  since not all features the network was trained to recognize may be strongly pronounced for a given image (instead mixtures of potential features may form), or because multiple features resemble each other at the given processing stage. Therefore, replacing a corrupted fmap with a non-corrupted fmap from a \textit{different} kernel can help to obtain an estimate of the original topology. We average all healthy (i.e., not containing out-of-bound activations) fmaps by
\begin{align}
&\find = \{i = 1 \ldots \Cout | \max(f_i) \leq \Tup, \min(f_i) \geq \Tlow \}, \notag \\ 
&\favg = \frac{1}{|\find|} \sum_{j \in \find} f_i.
\label{eq:favg}
\end{align}
If there are no healthy feature maps, $\favg$ will be the zero-tensor. 
%, and averaging will resemble clipping.
Subsequently, we replace $\oob$ values in a corrupted fmap with their counterparts from the estimate of Eq.~(\ref{eq:favg}),
\begin{equation}
r_{\text{favg}}(x) =
\begin{cases}
\favg(x) & \text{if $x > \Tup$ or $x < \Tlow$}, \\
x & \text{otherwise}.
\end{cases}
\label{eq:rfavg}
\end{equation}

\subsection{Results}

In Fig.~\ref{fig:sdc_all} we present results for the SDC mitigation experiments with different range supervision methods. Comparing $1$ and $10$ fault injections per input image, we note that the unprotected models are dramatically corrupted with an increasing fault rate (SDC rate becomes $\geq 0.50$ for weights, $\geq 0.32$ for neurons in the presence of $10$ faults). We can associate the SDC rate with the chance of unsuccessful mitigation, $1-P_{\text{mitigation}}$, in Eq.~\ref{eq:risk}.
Weight faults have a higher impact than neuron faults since they directly corrupt a multitude of activations in a layer's fmap output (in contrast to individual activations for neuron faults) and thus propagate faster than neuron faults.
%Further, the intrinsic robustness against faults is higher for the networks that are deeper in terms of layers. 

All the studied range restriction methods reduce the SDC rate by a significant margin, but perform differently for weight and neuron fault types:
For weight faults, we observe that \textit{Clipper}, \textit{Backflip}, and \textit{FmapAvg} are highly efficient in all three networks, with SDC rates suppressed to values of $\lesssim 0.01$ (SDC reduction of $>50\times$).
\textit{Ranger} provides a much weaker protection, in particular in the more shallow networks VGG-16 and AlexNet. 
\textit{FmapRescale} performs better than \textit{Ranger} but worse than the aforementioned methods.
The deepest studied network, ResNet-50, benefits the most from any type of range restriction in the presence of weight faults. 

When it comes to neuron faults (Fig.~\ref{fig:sdc_all}b), we see that \textit{Clipper} and \textit{Backflip} provide the best protection (SDC rate is suppressed to $<0.005$, reduction of $>38\times$), followed by the also very effective \textit{Ranger} (except for AlexNet).
\textit{FmapAvg} appears to be less efficient for higher fault rates in this scenario, while \textit{FmapRescale} again falls behind all the above.

Overall, we conclude that the pruning-inspired mitigation techniques \textit{Clipper} and \textit{Backflip} represent the best choices among the investigated ranger supervision methods, as they succeed in mitigating both weight and neuron faults to very small residual SDC rates.

In the experiments of Fig.~\ref{fig:sdc_all}, the encountered DUE rates for $1$ weight or neuron fault ($0.003$ for ResNet, $0.03$ for VGG-16 or AlexNet) are only slightly reduced by range restrictions.
However, for a fault rate of $10$ we find the following trends: \textbf{i)} For weights, the DUE is significantly reduced in ResNet (from $0.15$ to $0.002$), while rates in VGG ($0.22$) and AlexNet ($0.26$) remain.
\textbf{ii)} For neurons, \textit{Ranger}, \textit{Clipper} and \textit{Backflip} suppress the DUE rate by a factor of up to $2\times$ in all networks.
%on the other hand, ResNet has a lower DUE rate ($0.04$) and \textit{Ranger}, \textit{Clipper} and \textit{Backflip} can suppress the DUE rate by a factor of up to $2\times$ in all networks.

The studied range restriction techniques require different compute costs due to the different number of additional graph operations.
In PyTorch, not all needed functions can be implemented with the same efficiency though. For example, \textit{Ranger} is executed with a single \textit{clamp} operation, while no equivalent formulation is available for \textit{Clipper} and instead three operations are necessary (two masks to select oob values greater and smaller than the threshold, and a \textit{masked-fill} operation to clip to zero). 
As a consequence, measured latencies are framework-dependent and a fair comparison cannot be made at this point. 
Given the complexity of the protection operations, we may instead give a qualitative performance ranking of the described methods:
\textit{FmapRescale} appears to be the most expensive restriction method due to the needed number of operations, followed by \textit{FmapAvg} and \textit{Backflip}. \textit{Clipper} and \textit{Ranger} are the least complex, with the latter outperforming the former in the used framework, due to its more efficient use of optimized built-in operations.

%%%%%%%%%%%%%%%%%%%%%%%%%%%%%%%%%%%%%%%%%%%%%%%%%%%%%%%%%%%%%%%%%%%%%%%%%%%%%%%%

\begin{figure}[t]
\centering
\includegraphics[width=.5\textwidth]{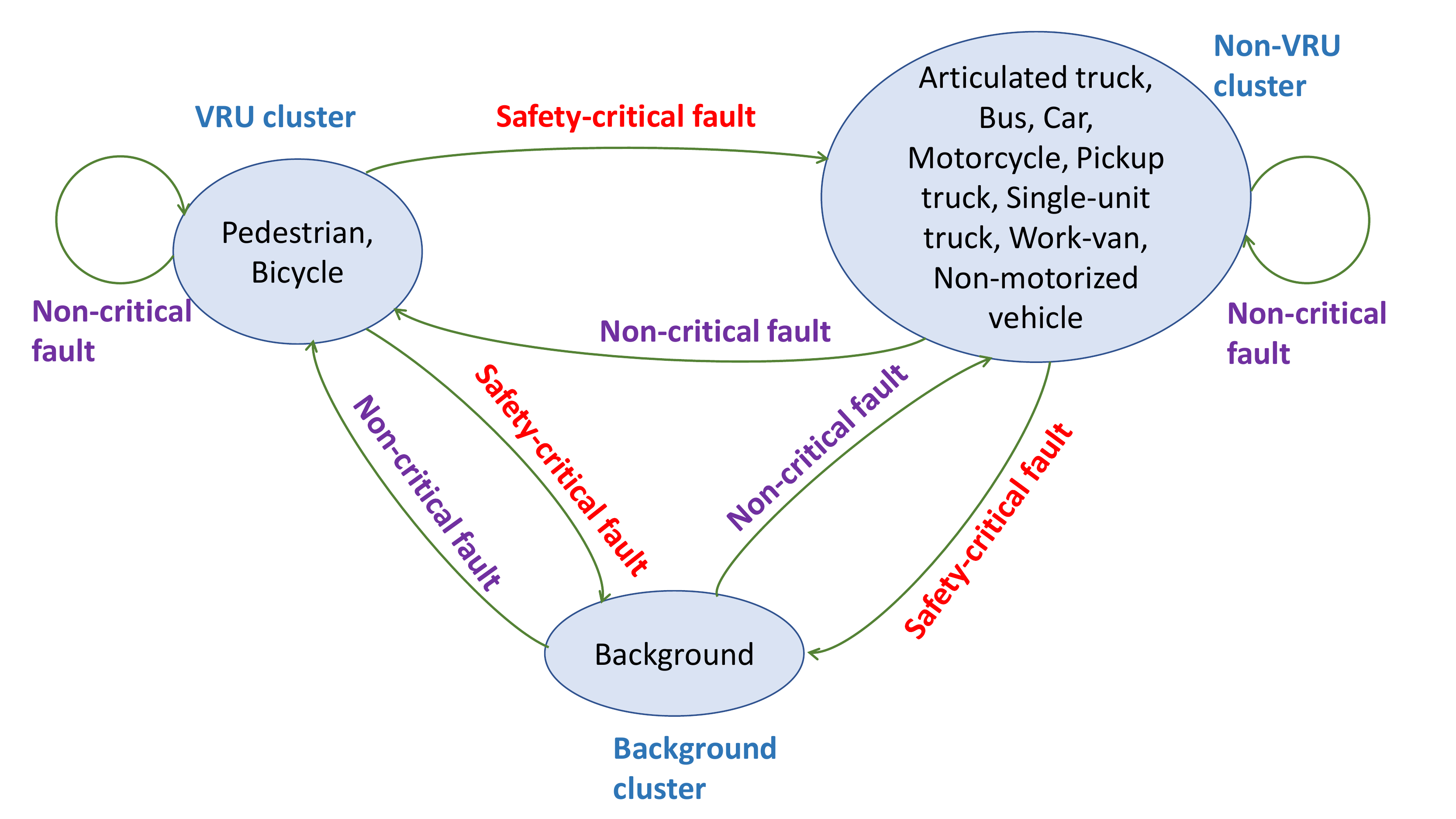} %[width=0.5\textwidth]
\caption{Formation of class clusters in MIOVision (VRU denotes vulnerable road user). We make the assumption here that confusions towards less vulnerable clusters are the most safety-critical ones.}
\label{fig:clusters}
\end{figure}

\section{Analysis of traffic camera use case}
\label{sec:safety}

As a selected safety-critical use case, we study object classification in the presence of soft errors with a retrained ResNet-50 and the MIOVision data set \cite{Luo2018}.
The data contains images of $11$ classes including for example pedestrian, bike, car, or background, that were taken by traffic cameras. The correct identification of an object type or category can be safety-critical for example to an automated vehicle that uses the support of infrastructure sensors for augmented perception \cite{Krammer2019}. 

However, not every class confusion is equally harmful. 
To estimate the severity of an error-induced misclassification we establish three clusters of vulnerable, as well as non-vulnerable road users (VRU or non-VRU), and background, see Fig.~\ref{fig:clusters}.
%The severity of a classifier prediction is then estimated in a binary way: 
Misclassifications that lead to the prediction of a class in a \textit{less vulnerable} cluster are assumed to be safety-critical ($\text{Severity}\approx 1$ in Eq.~\ref{eq:risk}, e.g., a pedestrian is misclassified as background), while confusions within the same cluster or towards a \textit{more vulnerable} cluster are considered non-critical ($\text{Severity}\approx 0$) as they typically lead only to similar or a more cautious behavior.
This binary estimation allows us quantify the overall risk as the portion of SDC events associated with the respective critical class confusions.
%For a conservative safety estimate, we study only weight faults in this section since they have a higher impact on the SDC.

\begin{figure}[!tbp]
     \centering
     \begin{subfigure}{0.41\textwidth}
         \centering
         \includegraphics[width=\textwidth]{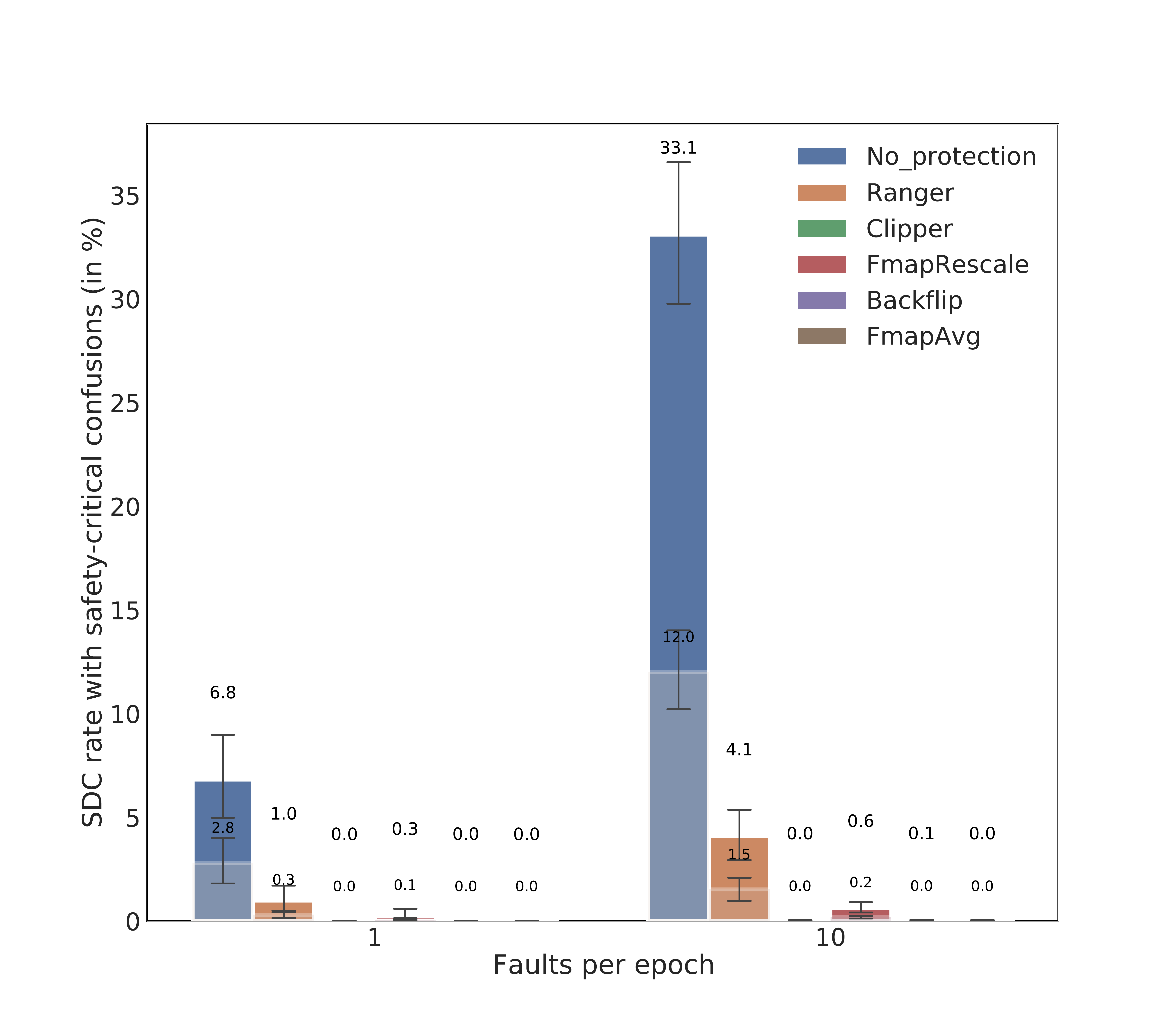}
         \caption{Weight faults}
         \label{fig:safety1}
     \end{subfigure} \\ %\hfill %\\
		%\vspace{-0.2cm}
     \begin{subfigure}{0.41\textwidth}
         \centering
         \includegraphics[width=\textwidth]{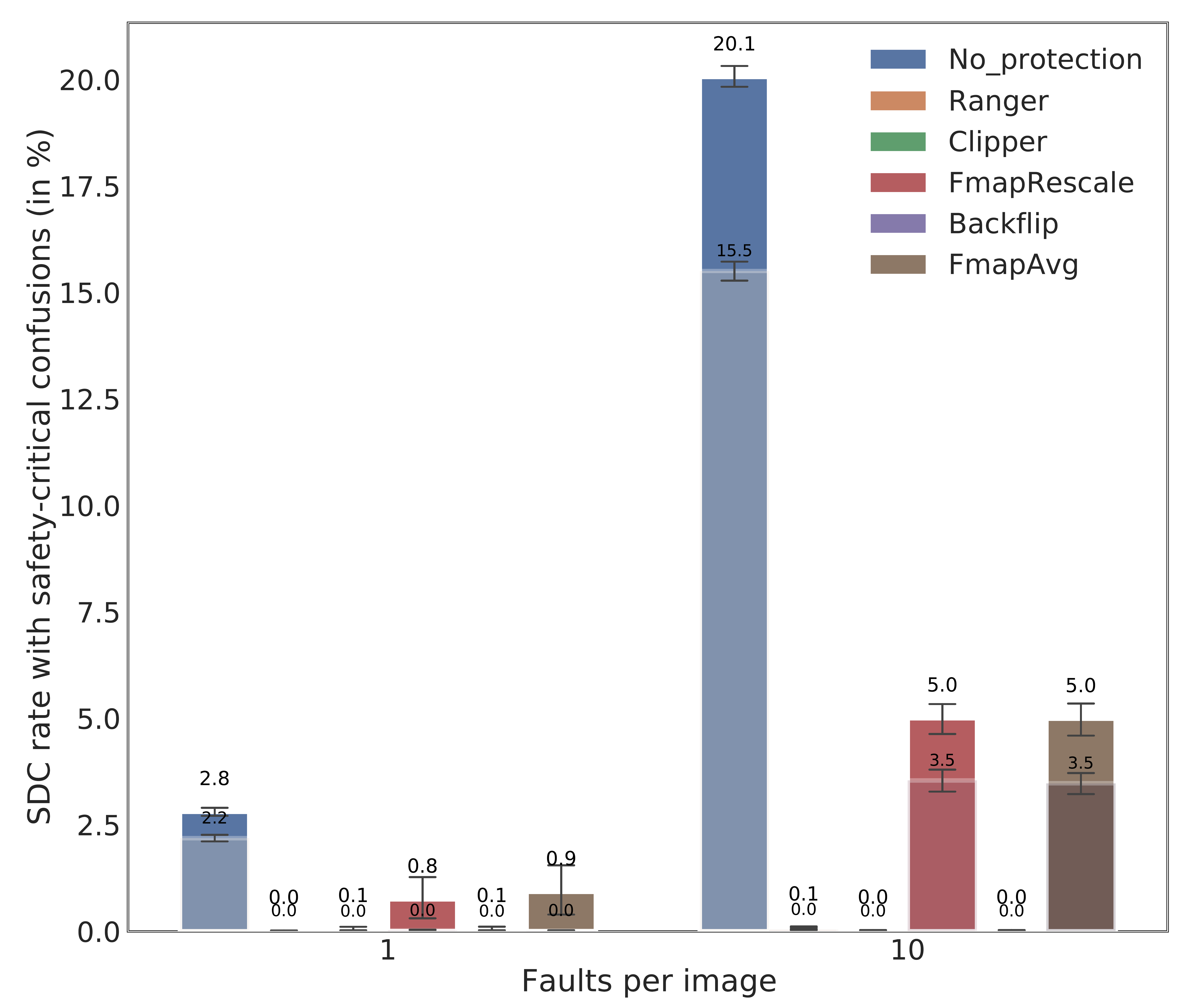}
         \caption{Neuron faults}
         \label{fig:safety2}
     \end{subfigure}
\caption{SDC rates for ResNet-50 and MIOvision. We inject $1$ and $10$ faults targeting bits $0-8$ in the network weights (a) and neurons (b).
The portion of safety-critical SDC events according to Fig.~\ref{fig:clusters} is displayed as a dark-color overlay.}
\label{fig:safety}
%\vspace{-0.3cm}
\end{figure}

From our results in Fig.~\ref{fig:safety} we make the following observations: 
\textbf{i)} The relative proportion of critical confusions is lower for weight than for neuron faults in the unprotected and most protected models. 
%This can be understood the following way: 
For weight faults, the most frequent confusions are from other classes to the class "car" (the most robust class of MIOVision, with the most images in the training set), which are statistically mostly non-critical.
Neuron faults, on the other hand, distort feature maps in a way that induces with the highest frequency misclassifications towards the class "background". Those events are all safety-critical (see Fig.~\ref{fig:clusters}), leading to a high critical-to-total SDC ratio.
%We attribute this behavior to the effect that weight faults propagate faster than neuron faults: With more corrupted values, the network loses more of its predictive capacity, so that incorrectly predicted classes are more randomly distributed. With more possibilities for non-critical ($84$) than for critical ($26$) confusions, see Fig.~\ref{fig:clusters}, the relative contribution of critical SDCs decreases and approaches a ratio of $26/84 \approx 0.31$.
%
%
\textbf{ii)} Range supervision is not only effective in reducing  the overall SDC count, but also suppresses the critical SDC count proportionally. 
For example, we observe that the most frequent critical class confusion caused by $1$ or $10$ weight faults is from the class "pedestrian" to "car" ($\approx 0.2$ of all critical SDC cases), where $>0.99$ of those cases can be mitigated by \textit{Clipper} or \textit{Backflip}. For neuron faults, the largest critical SDC contribution is from "pedestrian" to "background" ($1$ fault) or "car" to "background" ($10$ faults), both in about $0.1$ of all critical SDC cases. \textit{Clipper} or \textit{Backflip} are able to suppress $>0.91$ of those events.

As a consequence, all studied range-restricted models exhibit a critical-to-total SDC ratio that is similar to or lower than one of the unprotected network ($<0.41$ for weight, $<0.78$ for neuron faults), meaning that faults in the presence of range supervision have on average a similar or lower severity than faults that do not face range restrictions.
A lower ratio can be interpreted as a better preservation of the feature map topology: If the reconstructed features are more similar to the original features there is a higher chance of the incorrect class being similar to the original class and thus to stay within the same cluster. The total probability of critical SDC events -- and therefore the relative risk according to Eq.~\ref{eq:risk} -- is negligible in the studied setup in the presence of \textit{Clipper} or \textit{Backflip} range protection.

The mean DUE rates in the unprotected model are $0.0$ ($0.02$) for $1$ weight (neuron) fault and $0.11$ ($0.17$) for $10$ faults. Using any of the protection methods, the system's availability increases as DUE rates are negligible for $1$ fault, and reduce to $<0.03$ ($<0.05$) for $10$ weight (neuron) faults.

%%%%%%%%%%%%%%%%%%%%%%%%%%%%%%%%%%%%%%%%%%%%%%%%%%%%%%%%%%%%%%%%%%%%%%%%%%%%%%%%

\section{Conclusion}
\label{sec:conclusion}

In this paper, we investigated the efficacy of range supervision techniques for constructing a safety case for 
%We find that range supervision provides all the relevant elements to construct a safety case for 
computer vision AI applications that use Convolutional Neural Networks (CNNs) in the presence of platform soft errors.
In the given experimental setup, we demonstrated that the implementation of activation bounds allows for a highly efficient detection of SDC-inducing faults, most importantly featuring a recall of $>0.99$. Furthermore, we found that the range restriction layers can mitigate the once-detected faults effectively by mapping out-of-bound values back to the expected intervals. Exploring distinct restriction methods, we observed that \textit{Clipper} and \textit{Backflip} perform best for both weight and neuron faults and can reduce the residual SDC rate to $\lesssim 0.01$ (reduction by a factor of $>38\times$). Finally, we studied the selected use case of vehicle classification to quantify the impact of range restriction on the severity of SDC events (represented by cluster-wise class confusions). All discussed techniques reduce critical and non-critical events proportionally, meaning that the average severity of SDC is not increased. Therefore, we conclude that the presented approach reduces the overall risk and thus enhances the safety of the user in the presence of platform soft errors.

\section*{Acknowledgment}
Our research was partially funded by the Federal Ministry of Transport and Digital Infrastructure of Germany in the project Providentia++ (01MM19008). Further, this research was partially supported by a grant from the Natural Sciences and Engineering Research Council of Canada (NSERC), and a research gift from Intel to UBC.
%Funding was received from the European Union’s Horizon 2020 research and innovation program under grant agreement No. 956123.

%%%%%%%%%%%%%%%%%%%%%%%%%%%%%%%%%%%%%%%%%%%%%%%%%%%%%%%%%%%%%%%%%%%%%%%%%%%%%%%%
%%%%%%%%%%%%%%%%%%%%%%%%%%%%%%%%%%%%%%%%%%%%%%%%%%%%%%%%%%%%%%%%%%%%%%%%%%%%%%%%

\footnotesize
\bibliographystyle{IEEEtran} %IEEEtran
\bibliography{Bibliography_ML.bib}

%%%%%%%%%%%%%%%%%%%%%%%%%%%%%%%%%%%%%%%%%%%%%%%%%%%%%%%%%%%%%%%%%%%%%%%%%%%%%%%%
%\section*{APPENDIX}

\end{document}